\documentclass[twoside,11pt,letter]{article}
\input epsf
\usepackage{jair}
\usepackage{theapa}

\usepackage{amsmath, amssymb, graphicx,stmaryrd, color, framed, theorem,url}

\newcommand{\CalA}{{\cal A}}

\newcommand{\CalC}{{\cal C}}

\newcommand{\CalE}{{\cal E}}
\newcommand{\CalF}{{\cal F}}

\newcommand{\CalI}{{\cal I}}
\newcommand{\CalL}{{\cal L}}

\newcommand{\CalN}{{\cal N}}

\newcommand{\CalP}{{\cal P}}
\newcommand{\CalR}{{\cal R}}
\newcommand{\CalS}{{\cal S}}

\newcommand{\CalZ}{{\cal Z}}

\newcommand{\choice}{\mbox{\em choice\,}}

\newcommand{\prob}{\mbox{\em prob\,}}

\newcommand{\reward}{\mbox{\em reward\,}}

\newcommand{\Eff}{\mbox{\it Eff\,}}

\newcommand{\Pre}{\mbox{\it Pre\,}}

\newcommand{\forward}{\mbox{\it forward\,}}

\newcommand{\Ch}{\mbox{\it Ch\,}}

\newcommand{\bd}{\begin{document}}
\newcommand{\ed}{\end{document}}
\newcommand{\ba}{\begin{abstract}}
\newcommand{\ea}{\end{abstract}}
\newcommand{\br}{\begin{array}}
\newcommand{\er}{\end{array}}
\newcommand{\bfr}{\begin{flushright}}
\newcommand{\efr}{\end{flushright}}
\newcommand{\bfl}{\begin{flushleft}}
\newcommand{\efl}{\end{flushleft}}
\newcommand{\bb}{\begin{thebibliography}}
\newcommand{\eb}{\end{thebibliography}}
\newcommand{\bi}{\begin{itemize}}
\newcommand{\ei}{\end{itemize}}
\newcommand{\be}{\begin{enumerate}}
\newcommand{\ee}{\end{enumerate}}
\newcommand{\beq}{\begin{equation*}}
\newcommand{\eeq}{\end{equation*}}
\newcommand{\bq}{\begin{equation}}
\newcommand{\eq}{\end{equation}}
\newcommand{\bt}{\begin{tabbing}}
\newcommand{\et}{\end{tabbing}}
\newcommand{\bfig}{\begin{figure}}
\newcommand{\efig}{\end{figure}}
\newcommand{\bpic}{\begin{picture}}
\newcommand{\epic}{\end{picture}}
\newcommand{\bedef}{\begin{definition}}
\newcommand{\eedef}{\end{definition}}
\newcommand{\bsp}{\begin{slide*}}
\newcommand{\esp}{\end{slide*}}
\newcommand{\bsl}{\begin{slide}}
\newcommand{\esl}{\end{slide}}

\newtheorem{definition}{Definition}[section]

{\theorembodyfont{\rmfamily}  

\newtheorem{example}{Example}[section]}

\newtheorem{property}{Property}[section]

\newtheorem{remark}{Remark}[section]

\newtheorem{proposition}{Proposition}[section]

\newtheorem{theorem}{Theorem}[section]

\newtheorem{corollary}{Corollary}[section]

\newtheorem{lemma}{Lemma}[section]

\newcommand{\CN}{{\mbox{\rm\em CN\/}}}
\newcommand{\CNs}{{\mbox{\rm\em CN-state\/}}}
\newcommand{\CNss}{{\mbox{\rm\em CN-states\/}}}

\newcommand{\CND}{{\mbox{\rm\em CND\/}}}

\newcommand{\CNDs}{{\mbox{\rm\em CND-state\/}}}

\newcommand{\CNDss}{{\mbox{\rm\em CND-states\/}}}

\newcommand{\CalCs}{{\mbox{\rm\em C-state\/}}}

\newcommand{\CalCss}{{\mbox{\rm\em C-states\/}}}

\newcommand{\la}{\langle}
\newcommand{\ra}{\rangle}
\newcommand{\s}{\textsc{state}}
\newcommand{\act}{\textsc{action}}
\newcommand{\f}{\textsc{fluent}}
\newcommand{\lact}{\textsc{list-action}}

\newcommand{\LAO}{{\text{LAO}^*}}
\newcommand{\FOLAO}{{\text{FOLAO}^*}}
\newcommand{\FOVI}{{\text{FOVI}}}

\newcommand{\heading}[1]{

  \begin{center}

    \large{\sf{\textbf{#1}}}

  \end{center}

  \vspace{1ex minus 1ex}}

\newcommand{\If}{\mbox{\em if\,}}

\newcommand{\Locked}{\mbox{\em locked\,}}

\newcommand{\Cont}{\mbox{\em continue\,}}

\newcommand{\Alarm}{\mbox{\em alarm\,}}

\newcommand{\Green}{\mbox{\em green\,}}

\newcommand{\Red}{\mbox{\em red\,}}

\newcommand{\Current}{\mbox{\em current\,}}

\newcommand{\Next}{\mbox{\em next\,}}

\newcommand{\Drop}{\mbox{\em drop\,}}

\newcommand{\Floor}{\mbox{\em floor\,}}

\newcommand{\Fragile}{\mbox{\em fragile\,}}

\newcommand{\Broken}{\mbox{\em broken\,}}

\newcommand{\DropN}{\mbox{\em dropN\,}}

\newcommand{\DropF}{\mbox{\em dropF\,}}

\newcommand{\pickup}{\mbox{\em pickup\,}}

\newcommand{\pickupS}{\mbox{\em pickupS\,}}

\newcommand{\pickupF}{\mbox{\em pickupF\,}}

\newcommand{\putdownS}{\mbox{\em putdownS\,}}

\newcommand{\putdownF}{\mbox{\em putdownF\,}}

\newcommand{\IsRed}{\mbox{\em is-red\,}}

\newcommand{\IsBlue}{\mbox{\em is-blue\,}}

\newcommand{\IsGreen}{\mbox{\em is-green\,}}

\newcommand{\OnTopOf}{\mbox{\em on-top-of\,}}

\newcommand{\Holding}{\mbox{\em holding\,}}

\newcommand{\Top}{\mbox{\em Top\,}}

\newcommand{\Bottom}{\mbox{\em Bottom\,}}

\newcommand{\flucap}{\mbox{\sc FluCaP}}

\newenvironment{sketch}{\noindent\textbf{Sketch of the proof}}{\hfill $\square$}

\newenvironment{proof}{\noindent\textbf{Proof}}

\newsavebox{\fminibox}
\newlength{\fminilength}
\newenvironment{fminipage}[1][\linewidth]
{\setlength{\fminilength}{#1-2\fboxsep-2\fboxrule}%
   \begin{lrbox}{\fminibox}\begin{minipage}{\fminilength}}
{\end{minipage} \end{lrbox}\noindent\fbox{\usebox{\fminibox}}}

\newsavebox{\fmbox}
\newenvironment{fmpage}[1]
     {\begin{lrbox}{\fmbox}\begin{minipage}{#1}}
     {\end{minipage}\end{lrbox}\fbox{\usebox{\fmbox}}}

\definecolor{light}{gray}{.90}

\jairheading{27}{2006}{381-401}{12/05}{12/06}
\ShortHeadings{FluCaP: A Heuristic Search Planner for First-Order MDPs}{H\"olldobler, Karabaev \& Skvortsova}

\firstpageno{381}

\title{\flucap : A Heuristic Search Planner for First-Order MDPs}

\author{\name Steffen H\"olldobler \email sh@iccl.tu-dresden.de \\
       \name Eldar Karabaev \email eldar@iccl.tu-dresden.de \\
       \name Olga Skvortsova \email skvortsova@iccl.tu-dresden.de \\
        \addr  International Center for Computational Logic\\
               Technische Universit\"at Dresden, 
               Dresden, Germany }

\begin{document}

\engineeringnote

\maketitle


\begin{abstract}

We present a heuristic search algorithm for solving first-order Markov Decision Processes (FOMDPs). Our approach combines \emph{first-order state abstraction} that avoids evaluating states individually, and \emph{heuristic search} that avoids evaluating all states. Firstly, in contrast to existing systems, which start with propositionalizing the FOMDP and then perform state abstraction on its propositionalized version we apply state abstraction directly on the FOMDP avoiding propositionalization. This kind of abstraction is referred to as first-order state abstraction. Secondly, guided by an admissible heuristic, the search is restricted to those states that are reachable from the initial state. We demonstrate the usefulness of the above techniques for solving FOMDPs with a system, referred to as $\flucap$ (formerly, FCPlanner), that entered the probabilistic track of the 2004 International Planning Competition (IPC'2004) and demonstrated an advantage over other planners on the problems represented in first-order terms.

\end{abstract}

\section{Introduction}
\label{Introduction}

Markov decision processes (MDPs) have been adopted as a
representational and computational model for decision-theoretic
planning problems in much recent work, e.g., by~\citeA{barto}. The basic solution techniques for MDPs rely on the dynamic programming (DP) principle \cite{boutilierA}. Unfortunately, classical dynamic programming algorithms require explicit enumeration of the state space that grows exponentially with the number of variables relevant to the planning domain. Therefore, these algorithms do not scale up to complex AI planning problems.

However, several methods that avoid explicit state enumeration have been developed recently. One technique, referred to as state abstraction, exploits the structure of the factored MDP representation to solve problems efficiently, circumventing explicit state space enumeration~\cite{boutilierA}. Another technique, referred to as heuristic search, restricts the computation to states that are reachable from the initial state, e.g., RTDP by~\shortciteA{barto}, envelope DP by~\citeA{dean} and $\LAO$ by~\shortciteA{feng}. One existing approach that combines both these techniques is the symbolic $\LAO$ algorithm by~\shortciteA{feng} which performs heuristic search symbolically for factored MDPs. It exploits state abstraction, i.e., manipulates sets of states instead of individual states. More precisely, following the SPUDD approach by~\citeA{hoey}, all MDP components, value functions, policies, and admissible heuristic functions are compactly represented using algebraic decision diagrams (ADDs). This allows computations of the $\LAO$ algorithm to be performed efficiently using ADDs.

Following ideas of symbolic $\LAO$, given an initial state, we use an admissible heuristic to restrict search only to those states that are reachable from the initial state. Moreover, we exploit state abstraction in order to avoid evaluating states individually. Thus, our work is very much in the spirit of symbolic $\LAO$ but extends it in an important way. Whereas the symbolic $\LAO$ algorithm starts with propositionalization of the FOMDP, and only after that performs state abstraction on its propositionalized version by means of propositional ADDs, we apply state abstraction directly on the structure of the FOMDP, avoiding propositionalization. This kind of abstraction is referred to as \emph{first-order state abstraction}.

Recently, following work by~\citeA{boutilierB},~\shortciteA{hoelldoblerA} have developed an algorithm, referred to as first-order value iteration (FOVI) that exploits first-order state abstraction. The dynamics of an MDP is specified in the Probabilistic Fluent Calculus established by~\shortciteA{hoelldoblerB}, which is a first-order language for reasoning about states and actions. More precisely, FOVI produces a logical representation of value functions and policies by constructing first-order formulae that partition the state space into clusters, referred to as \emph{abstract states}. In effect, the algorithm performs value iteration on top of these clusters, obviating the need for explicit state enumeration. This allows problems that are represented in first-order terms to be solved without requiring explicit state enumeration or propositionalization. 

Indeed, propositionalizing FOMDPs can be very impractical: the number of propositions grows considerably with the number of domain objects and relations. This has a dramatic impact on the complexity of the algorithms that depends directly on the number of propositions. Finally, systems for solving FOMDPs that rely on propositionalizing states also propositionalize actions which is problematic in first-order domains, because the number of ground actions also grows dramatically with domain size.

In this paper, we address these limitations by proposing an approach for solving FOMDPs that combines first-order state abstraction and heuristic search in a novel way, exploiting the power of logical representations. Our algorithm can be viewed as a first-order generalization of $\LAO$, in which our contribution is to show how to perform heuristic search for first-order MDPs, circumventing their propositionalization. In fact, we show how to improve the performance of symbolic $\LAO$ by providing a compact first-order MDP representation using Probabilistic Fluent Calculus instead of propositional ADDs. Alternatively, our approach can be considered as a way to improve the efficiency of the FOVI algorithm by using heuristic search together with symbolic dynamic programming.

\section{First-order Representation of MDPs}

Recently, several representations for propositionally-factored MDPs have been proposed, including dynamic Bayesian networks by~\shortciteA{boutilierA} and ADDs by~\shortciteA{hoey}. For instance, the SPUDD algorithm by~\shortciteA{hoey} has been used to solve MDPs with hundreds of millions of states optimally, producing logical descriptions of value functions that involve only hundreds of distinct values. This work demonstrates that large MDPs, described in a logical fashion, can often be solved optimally by exploiting the logical structure of the problem.

Meanwhile, many realistic planning domains are best represented in first-order terms. However, most existing implemented solutions for first-order MDPs rely on propositionalization, i.e., eliminate all variables at the outset of a solution attempt by instantiating terms with all possible combinations of domain objects. This technique can be very impractical because the number of propositions grows dramatically with the number of domain objects and relations.

For example, consider the following goal statement taken from the colored Blocksworld scenario, where the blocks, in addition to unique identifiers, are associated with colors.
\[
\begin{array}{ll}
G=\exists X_0\ldots X_7.\ red(X_0)\wedge green(X_1)\wedge blue(X_2) \wedge red(X_3) \wedge red(X_4) \wedge \\
\hspace{28mm}  red(X_5) \wedge green(X_6) \wedge green(X_7) \wedge Tower(X_0,\ldots,X_7)~,
\end{array}
\]
where $Tower(X_0,\ldots,X_7)$ represents the fact that all eight blocks comprise one tower. We assume that the number of blocks in the domain and their color distribution agrees with that in the goal statement, namely there are eight blocks $a, b,\ldots, h$ in the domain, where four of them are red, three are green and one is blue. Then, the full propositionalization of the goal statement $G$ results in $4!3!1!=144$ different ground towers, because there are exactly that many ways of arranging four red, three green and one blue block in a tower of eight blocks with the required color characteristics.

The number of ground combinations, and hence, the complexity of reasoning in a propositional planner, depends dramatically on the number of blocks and, most importantly, on the number of colors in the domain. The fewer colors a domain contains, the harder it is to solve by a propositional planner. For example, a goal statement $G'$, that is the same as $G$ above, but all eight blocks are of the same color, results in $8!=40320$ ground towers, when grounded.

To address these limitations, we propose a concise representation of FOMDPs within the Probabilistic Fluent Calculus which is a logical approach to modelling dynamically changing systems based on first-order logic. But first, we briefly describe the basics of the theory of MDPs.

\subsection{MDPs}

A Markov decision process (MDP), is a tuple
$(\CalZ,\CalA,\CalP,\CalR, \CalC)$, where $\CalZ$ is a finite set of
states, $\CalA$ is a finite set of actions, and $\CalP: \CalZ
\times \CalZ \times \CalA \rightarrow [0,1]$, written $\CalP(z'|z,a)$, specifies transition
probabilities. In particular, $\CalP(z'|z,a)$ denotes the probability of ending up at state $z'$ given that the agent was in state $z$ and action $a$ was executed. $\CalR: \CalZ \rightarrow \mathbb{R}$ is a real-valued reward function associating with each state $z$ its immediate utility $\CalR(z)$. $\CalC: \CalA \rightarrow \mathbb{R}$ is a real-valued cost function associating a cost $\CalC(a)$ with each action $a$. A sequential decision problem consists of an MDP and is the
problem of finding a policy $\pi: \CalZ \rightarrow \CalA$ that
maximizes the total expected discounted reward received
when executing the policy $\pi$ over an infinite (or indefinite) horizon.

The value of state $z$, when starting in $z$ and following the policy $\pi$ afterwards, can be computed by the following system of linear equations:
\[
V_\pi(z) = \CalR(z) + \CalC(\pi(z))+\gamma \sum_{z' \in \CalZ}
\CalP(z'|z,\pi(z)) V_\pi(z'),
\]
where $0\leq\gamma\leq 1$ is a discount factor. We take $\gamma$ equal to 1 for indefinite-horizon problems only, i.e., when a goal is reached the system enters an absorbing state in which no further rewards or costs are accrued. The optimal value function $V^{*}$ satisfies:
\[
V^{*}(z)=\CalR(z) + \underset{a\in\CalA}{\text{max}}\{\CalC(a)+\gamma \sum_{z' \in \CalZ}\CalP(z'|z,a) V^{*}(z')\}~,
\]
for each $z\in\CalZ$.

For the competition, the expected total reward model was used as the optimality criterion. Without discounting, some care is required in the design of planning problems to ensure that the expected total reward is bounded for the optimal policy. The following restrictions were made for problems used in the planning competition:

\be
\item Each problem had a goal statement, identifying a set of absorbing goal states.
\item A positive reward was associated with transitioning into a goal state.

\item A cost was associated with each action.

\item A ``done" action was available in all states, which could be used to end further accumulation of reward.
\ee

These conditions ensure that an MDP model of a planning problem is a \emph{positive bounded model} described by~\shortciteA{puterman}. The only positive reward is for transitioning into a goal state. Since goal states are absorbing, that is, they have no outgoing transitions, the maximum value of any state is bounded by the goal reward. Furthermore, the ``done" action ensures that there is an action available in each state that guarantees a non-negative future reward.

\subsection{Probabilistic Fluent Calculus}

Fluent Calculus (FC) by~\shortciteA{hoelldoblerB} was originally set up as a first-order logic
program with equality using SLDE-resolution as the sole inference
rule. The Probabilistic Fluent Calculus (PFC) is an extension of the original FC for expressing planning domains with actions which have probabilistic effects.

\subsubsection*{States}

Formally, let $\Sigma$ denote a set of function symbols. We distinguish two function symbols in $\Sigma$, namely the binary function symbol $\circ$, which is associative, commutative, and admits the unit element, and a constant 1. Let $\Sigma_{-}=\Sigma\setminus\{\circ,1\}$. Non-variable $\Sigma_{-}$-terms are called \emph{fluents}. The function names of fluents are referred to as \emph{fluent names}. For example, $on(X,table)$ is a fluent meaning informally that some block $X$ is on the table, where $on$ is a fluent name. \emph{Fluent terms} are defined inductively as follows: 1 is a fluent term; each fluent is a fluent term; $F\circ G$ is a fluent term, if $F$ and $G$ are fluent terms. For example, $on(b,table)\circ holding(X)$ is a fluent term denoting informally that the block $b$ is on the table and some block $X$ is in the robot's gripper. In other words, freely occurring variables are assumed to be existentially quantified.

We assume that each fluent may occur at most once in a state. Moreover, function symbols, except for the binary $\circ$ operator, constant $1$, fluent names and constants, are disallowed. In addition, the binary function symbol $\circ$ is allowed to appear only as an outermost connective in a fluent term. We denote a set of fluents as $\CalF$ and a set of fluent terms as $\CalL^\CalF$, respectively. An \emph{abstract state} is defined by a pair $(P,\CalN)$, where $P\in \CalL^{\CalF}$ and $\CalN\subseteq \CalL^{\CalF}$. We denote individual states by $z$, $z_1$, $z_2$ etc.,\ abstract states by $Z$, $Z_1$, $Z_2$ etc. and a set of abstract states $\CalL_{PN}$.

\begin{figure}
\begin{minipage}{\linewidth}
    \includegraphics[scale=1.3]{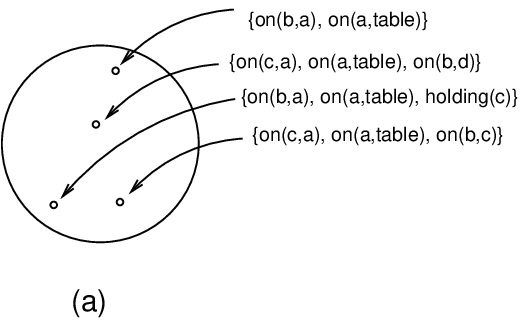}
    \hspace{3mm}
    \includegraphics[scale=1.3]{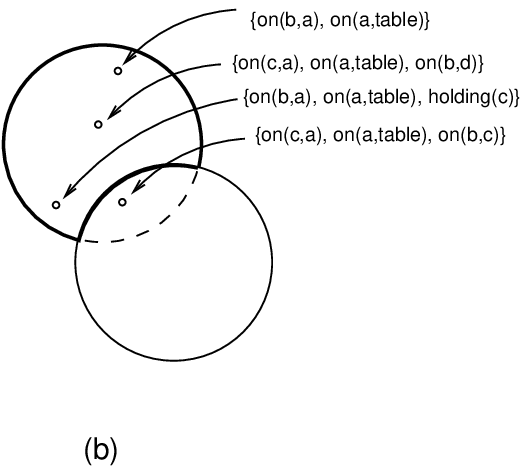}
    \begin{center}
      \includegraphics[scale=1.3]{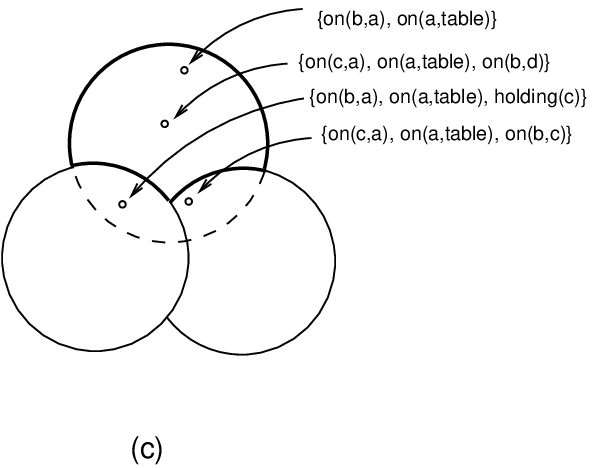}
    \end{center}
\end{minipage}
\caption{(a) Interpretation of the fluent term $F=on(X,a)\circ on(a,table)$; (b) Bold area is the interpretation of the abstract state $Z'=(on(X,a)\circ on(a,table),\{on(Y,X)\})$; (c) Bold area is the interpretation of the abstract state $Z=(on(X,a)\circ on(a,table),\{on(Y,X),holding(X')\})$.}
\label{fig1}
\end{figure}

The interpretation over $\CalF$, denoted as $\CalI$, is the pair $(\Delta,\cdot ^{\CalI})$, where the domain $\Delta$ is a set of all finite sets of ground fluents from $\CalF$; and an interpretation function $\cdot^{\CalI}$ which assigns to each fluent term $F$ a set $F^{\CalI}\subseteq \Delta$ and to each abstract state $Z=(P,\CalN)$ a set $Z^{\CalI}\subseteq \Delta$ as follows:
\begin{equation*}
\begin{array}{l}
F^{\CalI}=\{d\in \Delta \mid \exists \theta. F\theta \subseteq d\}\\
Z^{\CalI}=\{d\in \Delta \mid \exists \theta. P\theta \subseteq d\ \wedge \forall N\in \CalN.\ d\notin (N\theta)^{\CalI}\},
\end{array}
\label{eq:1}
\end{equation*}
where $\theta$ is a substitution. For example, Figure~\ref{fig1} depicts the interpretation of an abstract state $Z$
\[
Z=(on(X,a)\circ on(a,table),\{on(Y,X),holding(X')\})
\]
that can be informally read: There exists a block $X$ that is on the block $a$ which is on the table, there is no such block $Y$ that is on $X$ and there exists no such block $X'$ that the robot holds.
Since $Z^{\CalI}$ contains all such finite sets of ground fluents that satisfy the $P$-part and do not satisfy any of the elements of the $\CalN$-part, we subtract all sets of ground fluents that belong to each of $N_i\in\CalN$ from the set of ground fluents that correspond to the $P$-part. Thus, the bold area in Figure~\ref{fig1} contains exactly those sets of ground fluents (or, individual states) that do satisfy the $P$-part of $Z$ and none of the elements of its $\CalN$-part. For example, an individual state $ z_1=\{ on(b,a),on(a,table)\}$ belongs to $Z^{\CalI}$, whereas $ z_2=\{on(b,a),on(a,table),holding(c)\}$ does not. In other words, abstract states are characterized by means of conditions that must hold in each ground instance thereof and, thus, they
represent clusters of individual states. In this way, abstract states embody a form of state space abstraction. This kind of abstraction is referred to as \emph{first-order state abstraction}.

\subsubsection*{Actions}

\emph{Actions} are first-order terms starting with an action function symbol. For example, the action of picking up some block $X$ from another block $Y$ might be denoted as $\pickup(X,Y)$. Formally, let $N_a$ denote a set of action names disjoint with $\Sigma$. An \emph{action space} is a tuple $\CalA=(A,\Pre,\Eff$), where $A$ is a set of terms of the form $a(p_1,\ldots,p_n)$, referred to as \emph{actions}, with $a\in N_a$ and each $p_i$ being either a variable, or a constant; $\Pre:A\rightarrow\CalL_{PN}$ is a \emph{precondition} of $a$; and $\Eff:A\rightarrow\CalL_{PN}$ is an \emph{effect} of $a$.

So far, we have described deterministic actions only. But actions in PFC may have probabilistic effects as well. Similar to the work by~\shortciteA{boutilierB}, we decompose a stochastic action into deterministic primitives
under nature's control, referred to as \textit{nature's choices}. We use a relation symbol $\choice\!/2$ to model nature's choice. Consider the action $\pickup(X,Y)$:
\[
\begin{array}{l}
\choice(\pickup(X,Y),A) \leftrightarrow\ \\
\hspace{10mm}( A = \pickupS(X,Y)\vee A = \pickupF(X,Y) )~,
\end{array}
\]
where $\pickupS(X,Y)$ and $\pickupF(X,Y)$ define two nature's
choices for action $\pickup(X,Y)$, viz., that it succeeds or fails. For example, the nature's choice $\pickupS$ can be defined as follows:
\[
\begin{array}{l}
\Pre(\pickupS(X,Y)):=(on(X,Y)\circ e,\{on(W,X)\})\\
\Eff(\pickupS(X,Y)):=(holding(X),\{on(X,Y)\})~,
\end{array}
\]
where the fluent $e$ denotes the empty robot's gripper. For simplicity, we denote the set of nature's choices of an action $a$ as $\Ch(a):=\{a_j|\choice(a,a_j)\}$. Please note that nowhere do these action descriptions restrict the domain of discourse to some pre-specified set of blocks.

For each of nature's choices $a_j$ associated with an action $a$ we define the probability $\prob(a_j,a,Z)$ denoting the probability with which one of nature's choices $a_j$ is chosen in a state $Z$. For example,
\[
\begin{array}{l}
\prob(\pickupS(X,Y),\pickup(X,Y),Z) = .75
\end{array}
\]
states that the probability for the successful execution of the $\pickup$ action in state $Z$ is $.75$. 

In the next step, we define the reward function for
each state.  For example, we might want to give a reward of 500 to all states in which some block $X$ is on block $a$ and $0$, otherwise:
\[
\begin{array}{l}
  \reward(Z) = 500 \leftrightarrow Z\sqsubseteq(on(X,a),\emptyset) \\
  \reward(Z) = 0 \leftrightarrow Z\not\sqsubseteq(on(X,a),\emptyset)~,
\end{array}
\]
where $\sqsubseteq$ denotes the subsumption relation, which will be described in detail in Section~\ref{normalization}. One should observe that we have specified the reward function
without explicit state enumeration.  Instead, the state space is
divided into two abstract states depending on whether or not, a block $X$ is on block $a$.  Likewise, value functions can be specified with respect to the abstract states only.  This is in contrast to classical DP algorithms, in which the states are explicitly enumerated. Action costs can be analogously defined as follows: 
\[
cost(\pickup(X,Y))=3
\]
penalizing the execution of the $\pickup$-action with the value of 3.

\subsubsection*{Inference Mechanism}
\label{inference_mechanism}

Herein, we show how to perform inferences, i.e., compute successors of a given abstract state, with action schemata directly, avoiding unnecessary grounding. We note that computation of predecessors can be performed in a similar way.

Let $Z=(P,\CalN)$ be an abstract state, $a(p_1,\ldots, p_n)$ be an action with parameters $p_1,\ldots, p_n$, preconditions $\Pre(a)=(P_{p},\CalN_{p})$ and effects $\Eff(a)=(P_{e},\CalN_{e})$. Let $\theta$ and $\sigma$ be substitutions. An action $a(p_1,\ldots, p_n)$ is \emph{forward applicable}, or simply \emph{applicable}, to $Z$ with $\theta$ and $\sigma$, denoted as $\forward(Z,a,\theta,\sigma)$, if the following conditions hold:

\bi
\item[(f1)] $(P_{p}\circ U_1)\theta =_{AC1} P $
\item[(f2)] $\forall N_{p}\in \CalN_{p}.\exists N\in \CalN. (P\circ N \circ U_2)\sigma  =_{AC1}(P\circ N_{p})\theta~,$
\ei
where $U_1$ and $U_2$ are new AC1-variables and AC1 is the equational theory for $\circ$ that is represented by the following system of ``associativity'', ``commutativity'', and ``unit element'' equations:
\[
\begin{array}{ll}
\CalE_{AC1}=\{ & (\forall X,Y,Z)\ X\circ (Y \circ Z) = (X\circ Y)\circ Z\\
& (\forall X,Y)\ X\circ Y = Y\circ X\\
& (\forall X)\ X\circ 1 =X \hspace{40mm}\}~.
\end{array}
\]
 
In other words, the conditions (f1) and (f2) guarantee that $Z$ contains both positive and negative preconditions of the action $a$. If an action $a$ is forward applicable to $Z$ with $\theta$ and $\sigma$ then $Z_{succ}=(P',\CalN')$, where 
\begin{equation}
\begin{array}{l}
P':= (P_{e}\circ U_1)\theta\\
\CalN':=\CalN\sigma \setminus \CalN_{p}\theta\ \cup\  \CalN_{e}\theta
\end{array}
\label{eq:2}
\end{equation}
is referred to as the \emph{$a$-successor} of $Z$ with $\theta$ and $\sigma$ and denoted as $succ(Z,a,\theta,\sigma)$.

 For example, consider the action $\pickupS(X,Y)$ as defined above, take $Z=(P,\CalN)=(on(b,table)\circ on(X_1,b)\circ e,\{on(X_2,X_1)\})$. The action $\pickupS(X,Y)$ is forward applicable to $Z$ with $\theta=\{X\mapsto X_1, Y\mapsto b, U_1\mapsto on(b,table)\}$ and $\sigma=\{X_2 \mapsto W, U_2\mapsto 1\}$. Thus, $Z_{succ}=succ(Z,\pickupS(X,Y),\theta,\sigma)=(P',\CalN')$ with
\[
\begin{array}{l}
\hspace{-1mm} P'=holding(X_1)\circ on(b,table)\  \ \ \CalN'=\{on(X_1,b)\}~.
\end{array}
\]

\section{First-Order LAO*}

We present a generalization of the symbolic $\LAO$ algorithm by~\shortciteA{feng}, referred to as first-order $\LAO$ ($\FOLAO$), for solving FOMDPs. Symbolic $\LAO$ is a heuristic search algorithm that exploits state abstraction  for solving factored MDPs. Given an initial state, symbolic $\LAO$ uses an admissible heuristic to focus computation on the parts of the state space that are reachable from the initial state. Moreover, it specifies MDP components, value functions, policies, and admissible heuristics using propositional ADDs. This allows symbolic $\LAO$ to manipulate sets of states instead of individual states.

Despite the fact that symbolic $\LAO$ shows an advantageous behaviour in comparison to classical non-symbolic $\LAO$ by~\shortciteA{hansen} that evaluates states individually, it suffers from an important drawback. While solving FOMDPs, symbolic $\LAO$ propositionalizes the problem. This approach is impractical for large FOMDPs. Our intention is to show how to improve the performance of symbolic $\LAO$ by providing a compact first-order representation of MDPs so that the heuristic search can be performed without propositionalization. More precisely, we propose to switch the representational formalism for FOMDPs in symbolic $\LAO$ from propositional ADDs to Probabilistic Fluent Calculus. The $\FOLAO$ algorithm is presented in Figure~\ref{fig2}.

\begin{figure}[t]
\footnotesize
\begin{center}
\begin{fmpage}{\linewidth}
\begin{tabbing}
    {\tt} \= {\tt} \= {\tt} \= {\tt} \= {\tt} \= \kill
$policyExpansion(\pi,S^0,G)$\\
\>$E:=F:=\emptyset$\\
\>$from:=S^0$\\
\>\verb"repeat"\\
\>\>$to:=\underset{Z\in from}{\bigcup}\ \underset{a_j\in Ch(a)}{\bigcup}\{succ(Z,a_j,\theta,\sigma)\}$, \\
\>\> where $(a,\theta,\sigma):=\pi(Z)$\\
\>\>$F:=F\cup (to-G)$\\
\>\>$E:=E\cup from$\\
\>\>$from:=to \cap G -E$\\
\>\verb"until" $(from=\emptyset)$\\
\>$E:=E\cup F$\\
\>$G:=G\cup F$\\
\>\verb"return" $(E,F,G)$\\\\

$\FOVI(E,\CalA, prob, reward, cost, \gamma, V)$\\

\> \verb"repeat"\\
\>\> $V':=V$\\
\>\> \verb"loop" for each $Z\in E$\\
\>\>\> \verb"loop" for each $a\in\CalA$\\
\>\>\>\> \verb"loop" for each $\theta$, $\sigma$ such that $\forward(Z,a,\theta,\sigma)$\\
\>\>\>\>\> $Q(Z,a,\theta,\sigma):=reward(Z)+cost(a)+$\\
\>\>\>\>\> \hspace{5mm}$\gamma \underset{a_j\in Ch(a)}{\sum}prob(a_j,a,Z)\cdot V'(succ(Z,a_j,\theta,\sigma))$\\
\>\>\>\> \verb"end loop"\\
\>\>\> \verb"end loop"\\
\>\>\> $V(Z):=\underset{(a,\theta,\sigma)}{\text{max}}\ Q(Z,a,\theta,\sigma)$\\
\>\>\verb"end loop"\\
\>\> $V:=normalize(V)$\\
\>\> $r:=\|V-V'\|$\\
\>\verb"until" stopping criterion\\
\> $\pi:=extractPolicy(V)$\\
\> \verb"return" $(V,\pi,r)$\\\\

$\FOLAO(\CalA, prob, reward, cost, \gamma, S^0, h, \varepsilon)$\\

\>$V:=h$\\
\>$G:=\emptyset$\\
\>For each $Z\in S^0$, initialize $\pi$ with an arbitrary action\\
\> \verb"repeat"\\
\>\>\>$(E,F,G):=policyExpansion(\pi,S^0,G)$\\
\>\>\>$(V,\pi,r):=\FOVI(E,\CalA, prob, reward, cost, \gamma, V)$\\
\>\verb"until" $(F=\emptyset)$ \verb"and" $r\leq \varepsilon$\\
\>\verb"return" $(\pi,V)$

\end{tabbing}
\end{fmpage}
\end{center}
\caption{First-order $\LAO$ algorithm.}
\label{fig2}
\end{figure}

As symbolic $\LAO$, $\FOLAO$ has two phases that alternate until a complete solution is found, which is guaranteed to be optimal. First, it expands the best partial policy and evaluates the states on its fringe using an admissible heuristic function. Then it performs dynamic programming on the states visited by the best partial policy, to update their values and possibly revise the current best partial policy. We note that we focus on partial policies that map a subcollection of states into actions.

In the policy expansion step, we perform reachability analysis to find the set $F$ of states that have not yet been expanded, but are reachable from the set $S^0$ of initial states by following the partial policy $\pi$. The set of states $G$ contains states that have been expanded so far. By expanding a partial policy we mean that it will be defined for a larger set of states in the dynamic programming step. In symbolic $\LAO$, reachability analysis on ADDs is performed by means of the \emph{image} operator from symbolic model checking, that computes the set $to$ of successor states following the best current policy. Instead, in $\FOLAO$, we apply the $succ$-operator, defined in Equation~\ref{eq:2}. One should observe that since the reachability analysis in $\FOLAO$ is performed on abstract states that are defined as first-order entities, the reasoning about successor states is kept on the first-order level. In contrast, symbolic $\LAO$ would first instantiate $S^0$ with all possible combinations of objects, in order to be able to perform computations using propositional ADDs later on.

In contrast to symbolic $\LAO$, where the dynamic programming step is performed using a modified version of SPUDD, we employ a modified first-order value iteration algorithm ($\FOVI$). The original $\FOVI$ by~\shortciteA{hoelldoblerA} performs value iteration over the entire state space. We modify it so that it computes on states that are reachable from the initial states, more precisely, on the set $E$ of states that are visited by the best current partial policy. In this way, we improve the efficiency of the original $\FOVI$ algorithm by using reachability analysis together with symbolic dynamic programming. $\FOVI$ produces a PFC representation of value functions and policies by constructing first-order formulae that partition the state space into abstract states. In effect, it performs value iteration on top of abstract states, obviating the need for explicit state enumeration. 

Given a FOMDP and a value function represented in PFC, $\FOVI$ returns the best partial value function $V$, the best partial policy $\pi$ and the residual $r$. In order to update the values of the states $Z$ in $E$, we assign the values from the current value function to the successors of $Z$. We compute successors with respect to all nature's choices $a_j$. The residual $r$ is computed as the absolute value of the largest difference between the current and the newly computed value functions $V'$ and $V$, respectively. We note that the newly computed value function $V$ is taken in its normalized form, i.e., as a result of the $normalize$ procedure that will be described in Section~\ref{normalization}. Extraction of a best partial policy $\pi$ is straightforward: One simply needs to extract the maximizing actions from the best partial value function $V$.

As with symbolic $\LAO$, $\FOLAO$ converges to an $\varepsilon$-optimal policy when three conditions are met: (1) its current policy does not have any unexpanded states, (2) the residual $r$ is less than the predefined threshold $\varepsilon$, and (3) the value function is initialized with an admissible heuristic. The original convergence proofs for $\LAO$ and symbolic $\LAO$ by~\shortciteA{hansen} carry over in a straightforward way to $\FOLAO$.

When calling $\FOLAO$, we initialize the value function with an admissible heuristic function $h$ that focuses the search on a subset of reachable states. A simple way to create an admissible heuristic is to use dynamic programming to compute an approximate value function. Therefore, in order to obtain an admissible heuristic $h$ in $\FOLAO$, we perform several iterations of the original $\FOVI$. We start the algorithm on an initial value function that is admissible. Since each step of $\FOVI$ preserves admissibility, the resulting value function is admissible as well. The initial value function assigns the goal reward to each state thereby overestimating the optimal value, since the goal reward is the maximal possible reward.

Since all computations of $\FOLAO$ are performed on abstract states instead of individual states, FOMDPs are solved avoiding explicit state and action enumeration and propositionalization. The first-order reasoning leads to better performance of $\FOLAO$ in comparison to symbolic $\LAO$, as shown in Section~\ref{evaluation}.

\subsection{Policy Expansion}

\begin{figure}
 \includegraphics[scale=0.8]{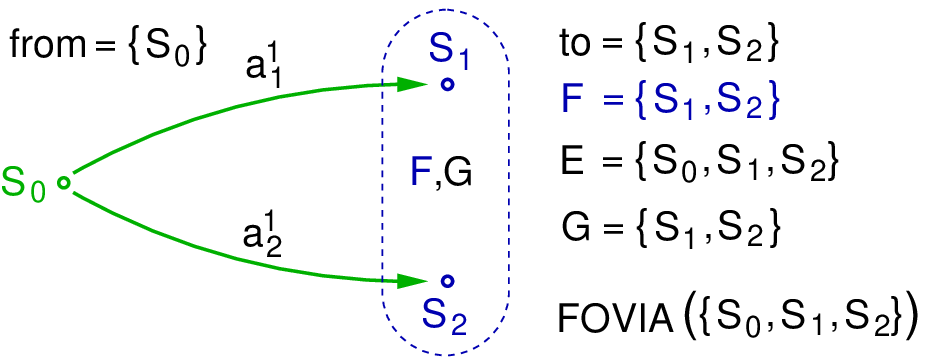}
 \includegraphics[scale=0.8]{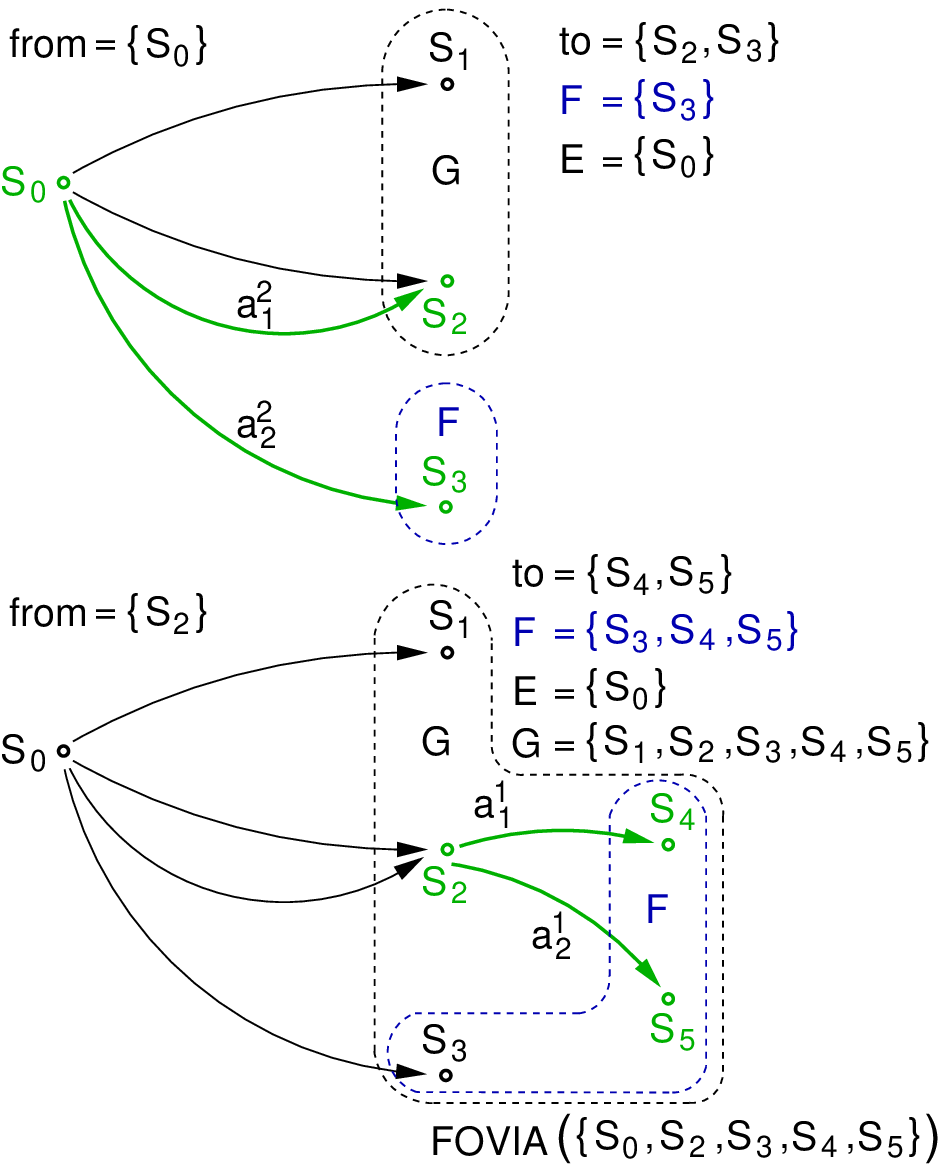}

\caption{Policy Expansion.}
\label{fig3}
\end{figure}

The policy expansion step in $\FOLAO$ is very similar to the one in the symbolic $\LAO$ algorithm. Therefore, we illustrate the expansion procedure by means of an example. Assume that we start from the initial state $Z_0$ and two nondeterministic actions $a^1$ and $a^2$ are applicable in $Z_0$, each having two outcomes $a^1_1$, $a^1_2$ and $a^2_1$, $a^2_2$, respectively. Without loss of generality, we assume that the current best policy $\pi$ chooses $a^1$ as an optimal action at state $Z_0$. We construct the successors $Z_1$ and $Z_2$ of $Z_0$ with respect to both outcomes $a^1_1$ and $a^1_2$ of the action $a^1$.

The fringe set $F$ as well as the set $G$ of states expanded so far
contain the states $Z_1$ and $Z_2$ only, whereas, the set $E$ of states
visited by the best current partial policy gets the state $Z_0$ in
addition. See Figure~\ref{fig3}a. In the next step, $\FOVI$ is performed on the set $E$. We assume that
the values have been updated in such a way that $a^2$ becomes an optimal action in $Z_0$. Thus, the successors of $Z_0$ have to be recomputed with respect to the optimal action $a^2$. See Figure~\ref{fig3}b.

One should observe that one of the $a^2$-successors of $Z_0$, namely $Z_2$, is an element of the set $G$ and thus, it has been contained already in the fringe $F$ during the previous expansion step. Hence, the state $Z_2$ should be expanded and its value recomputed. This is shown in Figure~\ref{fig3}c, where states $Z_4$ and $Z_5$ are $a^1$-successors of $Z_2$, under assumption that $a^1$ is an optimal action in $Z_2$. As a result, the fringe set $F$ contains the newly discovered states $Z_3$, $Z_4$ and $Z_5$ and we perform $\FOVI$ on $E=\{Z_0,Z_2,Z_3,Z_4,Z_5\}$. The state $Z_1$ is not contained in $E$, because it does not belong to the best current partial policy, and the dynamic programming step is performed only on the states that were visited by the best current partial policy.

\subsection{First-Order Value Iteration}
\label{fovia}

In $\FOLAO$, the first-order value iteration algorithm ($\FOVI$) serves two purposes: First, we perform several iterations of $\FOVI$ in order to create an admissible heuristic $h$ in $\FOLAO$. Second, in the dynamic programming step of $\FOLAO$, we apply $\FOVI$ on the states visited by the best partial policy in order to update their values and possibly revise the current best partial policy.

The original $\FOVI$ by~\shortciteA{hoelldoblerA} takes a finite state space of abstract states, a finite set of stochastic actions, real-valued reward and cost functions, and an initial value function as input. It produces a first-order representation of the optimal value function and policy by exploiting the logical structure of a FOMDP. Thus, FOVI can be seen as a first-order counterpart of the classical value iteration algorithm by~\shortciteA{bellman}.

\subsubsection{Normalization}
\label{normalization}

Following the ideas of~\shortciteA{boutilierB}, FOVI relies on the normalization of the state space that represents the value function. By normalization of a state space, we mean an equivalence-preserving  procedure that reduces the size of a state space. This would have an effect only if a state space contains redundant entries, which is usually the case in symbolic computations.

Although normalization is considered to be an important issue, it has been done by hand so far. To the best of our knowledge, the preliminary implementation of the approach by~\shortciteA{boutilierB} performs only rudimentary logical simplifications and the authors suggest using an automated first-order theorem prover for the normalization task. ~\shortciteA{hoelldoblerA} have developed an automated normalization procedure for FOVI that, given a state space, delivers an equivalent one that contains no redundancy. The technique employs the notion of a subsumption relation.

More formally, let $Z_1=(P_1,\CalN_1)$ and $Z_2=(P_2,\CalN_2)$ be abstract states. Then $Z_1$ is said to be {\em subsumed\/} by $Z_2$, written $Z_1\sqsubseteq Z_2$, if and only if there exist substitutions $\theta$ and $\sigma$ such that the following conditions hold:
\bi
\item[(s1)] $(P_2\circ U_1)\theta =_{AC1} P_1 $
\item[(s2)] $\forall N_2\in \CalN_2. \exists N_1\in \CalN_1. (P_1\circ N_1\circ U_2)\sigma=_{AC1} (P_1\circ N_2)\theta$~,
\ei
where $U_1$ and $U_2$ are new AC1-variables. The motivation for the notion of subsumption on abstract states is inherited from the notion of $\theta$-subsumption between first-order clauses by~\shortciteA{robinson} with the difference that abstract states contain more complicated negative parts in contrast to the first-order clauses.

For example, consider two abstract states $Z_1$ and $Z_2$ that are defined as follows:
\[
\begin{array}{l}
Z_1=(on(X_1,a)\circ on(a,table),\{red(Y_1)\})\\
Z_2=(on(X_2,a),\{red(X_2)\})~,
\end{array}
\]
where $Z_1$ informally asserts that some block $X_1$ is on the block $a$ which is on the table and no blocks are red. Whereas $Z_2$ informally states that some block $X_2$ is on the block $a$ and $X_2$ is not red. We show that $Z_1\sqsubseteq Z_2$. The relation holds since both conditions (s1) and (s2) are satisfied. Indeed,
\[
(on(X_2,a)\circ U_1)\theta =_{AC1} on(X_1,a)\circ on(a,table)
\]
and
\[
\begin{array}{l}
(on(X_1,a)\circ on(a,table)\circ red(Y_1)\circ U_2)\sigma = (on(X_1,a)\circ on(a,table)\circ red(X_2))\theta
\end{array}
\]
with $\theta=\{X_2\mapsto X_1, U_1\mapsto on(a,table)\}$ and $\sigma=\{Y_1\mapsto X_1, U_2\mapsto 1\}$.

One should note that subsumption in the language of abstract states inherits the complexity bounds of $\theta$-subsumption~\shortcite{kapur}. Namely, deciding subsumption between two abstract states is NP-complete, in general. However,~\shortciteA{karabaev} have recently developed an efficient algorithm that delivers all solutions of the subsumption problem for the case where abstract states are fluent terms.

For the purpose of normalization, it is convenient to represent the value function as a set of pairs of the form $\la Z,\alpha\ra$, where $Z$ is an abstract state and $\alpha$ is a real value. In essence, the normalization algorithm can be seen as an exhaustive application of the following simplification rule to the value function $V$.
\[ 
\Large \frac{\la Z_1, \alpha \ra\ \ \ \la Z_2, \alpha \ra}{\langle Z_2,\alpha \rangle}\ Z_1 \sqsubseteq Z_2
  \normalsize
\]
Table~\ref{t:1} illustrates the importance of the normalization algorithm by providing some representative timing results for the first ten iterations of FOVI. The experiments were carried out on the problem taken from the colored Blocksworld scenario consisting of ten blocks. Even on such a relatively simple problem FOVI with the normalization switched off does not scale beyond the sixth iteration.

\begin{table}[t]
  \footnotesize
\begin{center}
  \begin{tabular}{|c||c|c||c|c||c||c|}
   \hline
\textsf{N} &
\multicolumn{2}{c||}{\textsf{Number of states}} &
\multicolumn{2}{c||}{\textsf{Time, msec}} &
\textsf{Runtime, msec} &
\textsf{Runtime w/o norm, msec} \\ \cline{2-5}

 & $\CalS_{\textsf{update}}$& $\CalS_{\textsf{norm}}$ & \textsf{Update} & \textsf{Norm}  & & \\
\hline \hline
0  &    9     & 6    & 144        & 1     & 145 & 144  \\
1  &   24     & 14    & 393       & 3     & 396    & 593\\
2  &   94     & 23    & 884       & 12    & 896    & 2219\\
3  &   129     & 33    & 1377     & 16    & 1393    & 13293\\
4  &   328     & 39    & 2079     & 46    & 2125    & 77514 \\
5  &   361     & 48    & 2519     & 51    & 2570    & 805753\\
6  &   604     & 52    & 3268     & 107   & 3375    & n/a   \\
7  &   627     & 54    & 3534     & 110   & 3644    & n/a\\
8  &   795     & 56    & 3873     & 157   & 4030    & n/a\\
9  &   811     & 59    & 4131     & 154   & 4285    & n/a\\
\hline
 
  \end{tabular}
\end{center}
  \caption{Representative timing results for first ten iterations of FOVI.}
  \label{t:1}
\end{table}

The results in Table~\ref{t:1} demonstrate that the normalization during some iteration of FOVI dramatically shrinks the computational effort during the next iterations. The columns labelled $\CalS_{\textsf{update}}$ and $\CalS_{\textsf{norm}}$ show the size of the state space after performing the value updates and the normalization, respectively. For example, the normalization factor, i.e., the ratio between the number $\CalS_{\textsf{update}}$ of states obtained after performing one update step and the number $\CalS_{\textsf{norm}}$ of states obtained after performing the normalization step, at the seventh iteration is 11.6. This means that more than ninety percent of the state space contained redundant information. The fourth and fifth columns in Table~\ref{t:1} contain the time \textsf{Update} and \textsf{Norm} spent on performing value updates and on the normalization, respectively. The total runtime \textsf{Runtime}, when the normalization is switched on, is given in the sixth column. The seventh column labelled \textsf{Runtime w/o norm} depicts the total runtime of FOVI when the normalization is switched off. If we would sum up all values in the seventh column and the values in the sixth column up to the sixth iteration inclusively, subtract the latter from the former and divide the result by the total time \textsf{Norm} needed for performing normalization during the first six iterations, then we would obtain the normalization gain of about three orders of magnitude.

\section{Experimental Evaluation}
\label{evaluation}

We demonstrate the advantages of combining the heuristic search together with first-order state abstraction on a system, referred to as $\flucap$, that has successfully entered the probabilistic track of the 2004 International Planning Competition (IPC'2004). The experimental results were all obtained using RedHat Linux running on a 3.4GHz Pentium IV machine with 3GB of RAM.

In Table~\ref{t:2}, we present the performance comparison of $\flucap$ together with symbolic $\LAO$ on examples taken from the colored Blocksworld (BW) scenario that was introduced during IPC'2004.

Our main objective was to investigate whether first-order state abstraction using logic could improve the computational behaviour of a planning system for solving FOMDPs. The colored BW problems were our main interest since they were the only ones represented in first-order terms and hence the only ones that allowed us to make use of the first-order state abstraction. Therefore, we have concentrated on the design of a domain-dependent planning system that was tuned for the problems taken from the Blocksworld scenario.

The colored BW problems differ from the classical BW ones in that, along with the unique identifier, each block is assigned a specific color. A goal formula, specified in first-order terms, provides an arrangement of colors instead of an arrangement of blocks.

\begin{table*}[tb]

  \footnotesize
\begin{center}  
\begin{tabular}{|c|c||c|c|c|c||c|c|c|c||c|c||c|c|c||c|c||c|}
   \hline
\multicolumn{2}{|c||}{\hspace{-2mm}\textsf{Problem}\hspace{-5mm}} &
\multicolumn{4}{c||}{\hspace{-1mm}\textsf{Total av. reward}\hspace{-2mm}} &
\multicolumn{4}{c||}{\hspace{-2mm}\textsf{Total time, sec.}\hspace{-2mm}}&
\multicolumn{2}{c||}{\hspace{-2mm}\textsf{H.time, sec.}\hspace{-2mm}}&
\multicolumn{3}{c||}{\hspace{-2mm}\textsf{NAS}\hspace{-2mm}}&
\multicolumn{2}{c|}{\hspace{-2mm}\textsf{NGS}, $\times \textsf{10}^\textsf{3}$\hspace{-2mm}}&
\hspace{-2mm}\textsf{\%}\hspace{-2mm}\\ \cline{1-18}
\hspace{-2mm}\textsf{B}\hspace{-2mm}&\hspace{-2mm}\textsf{C}\hspace{-2mm}&\hspace{-2mm}\rotatebox{90}{\textsf{LAO*}}\hspace{-2mm}&\hspace{-2mm}\rotatebox{90}{\textsf{FluCaP}}\hspace{-2mm}&\hspace{-2mm}\rotatebox{90}{\textsf{FOVI}}\hspace{-2mm}&\hspace{-2mm}\rotatebox{90}{\textsf{FluCaP}$^\textsf{--}$}\hspace{-2mm}&\hspace{-2mm}\rotatebox{90}{\textsf{LAO*}}\hspace{-2mm}&\hspace{-2mm}\rotatebox{90}{\textsf{FluCaP}}\hspace{-2mm}&\hspace{-2mm}\rotatebox{90}{\textsf{FOVI}}\hspace{-2mm}&\hspace{-2mm}\rotatebox{90}{\textsf{FluCaP}$^\textsf{--}$}\hspace{-2mm}&\hspace{-2mm}\rotatebox{90}{\textsf{LAO*}}\hspace{-2mm}&\hspace{-2mm}\rotatebox{90}{\textsf{FluCaP}}\hspace{-2mm}&\hspace{-2mm}\rotatebox{90}{\textsf{LAO*}}\hspace{-2mm}&\hspace{-2mm} \rotatebox{90}{\textsf{FluCaP}}\hspace{-2mm}& \hspace{-2mm} \rotatebox{90}{\textsf{FOVI}}\hspace{-2mm}&\hspace{-2mm}\rotatebox{90}{\textsf{LAO*}}\hspace{-2mm}&\hspace{-2mm} \rotatebox{90}{\textsf{FluCaP}}\hspace{-2mm} & \hspace{-2mm} \rotatebox{90}{\textsf{FluCaP}}
\\
\hline \hline

  \hspace{-2mm}&\hspace{-2mm}4\hspace{-2mm}&\hspace{-2mm}494\hspace{-2mm}&\hspace{-2mm}494\hspace{-2mm}&\hspace{-2mm}494\hspace{-2mm}&\hspace{-2mm}494\hspace{-2mm}&\hspace{-2mm}22.3\hspace{-2mm}&\hspace{-2mm}22.0\hspace{-2mm}&\hspace{-2mm}23.4\hspace{-2mm}&\hspace{-2mm}31.1\hspace{-2mm}&\hspace{-2mm}8.7\hspace{-2mm}&\hspace{-2mm}4.2\hspace{-2mm}&\hspace{-2mm}35\hspace{-2mm}&\hspace{-2mm}410\hspace{-2mm}&\hspace{-5mm} 1077\hspace{-2mm} &\hspace{-2mm}0.86\hspace{-2mm}&\hspace{-2mm}0.82 \hspace{-2mm}& \hspace{-2mm}2.7\hspace{-2mm}\\

\hspace{-2mm}5\hspace{-2mm}&\hspace{-2mm}3\hspace{-2mm}&\hspace{-2mm}496\hspace{-2mm}&\hspace{-2mm}495\hspace{-2mm}&\hspace{-2mm}495\hspace{-2mm}&\hspace{-2mm}496\hspace{-2mm}&\hspace{-2mm}23.1\hspace{-2mm}&\hspace{-2mm}17.8\hspace{-2mm}&\hspace{-2mm}22.7\hspace{-2mm}&\hspace{-2mm}25.1\hspace{-2mm}&\hspace{-2mm}9.5\hspace{-2mm}&\hspace{-2mm}1.3\hspace{-2mm}&\hspace{-2mm}34\hspace{-2mm}&\hspace{-2mm}172\hspace{-2mm}& \hspace{-2mm}687 \hspace{-2mm}& \hspace{-2mm}0.86\hspace{-2mm}&\hspace{-4mm}0.68\hspace{-2mm} & \hspace{-2mm}2.1\hspace{-2mm}\\

  \hspace{-2mm}&\hspace{-2mm}2\hspace{-2mm}&\hspace{-2mm}496\hspace{-2mm}&\hspace{-2mm}495\hspace{-2mm}&\hspace{-2mm}495\hspace{-2mm}&\hspace{-2mm}495\hspace{-2mm}&\hspace{-2mm}27.3\hspace{-2mm}&\hspace{-2mm}11.7\hspace{-2mm}&\hspace{-2mm}15.7\hspace{-2mm}&\hspace{-2mm}16.5\hspace{-2mm}&\hspace{-2mm}12.7\hspace{-2mm}&\hspace{-2mm}0.3\hspace{-2mm}&\hspace{-2mm}32\hspace{-2mm}&\hspace{-2mm}55\hspace{-2mm}& \hspace{-3mm} 278 \hspace{-2mm}& \hspace{-2mm}0.86\hspace{-2mm}&\hspace{-2mm}0.66 \hspace{-2mm}& \hspace{-2mm}1.9\hspace{-2mm}\\

\hline \hline

  \hspace{-2mm}&\hspace{-2mm}4\hspace{-2mm}&\hspace{-2mm}493\hspace{-2mm}&\hspace{-2mm}493\hspace{-2mm}&\hspace{-2mm}493\hspace{-2mm}&\hspace{-2mm}493\hspace{-2mm}&\hspace{-2mm}137.6\hspace{-2mm}&\hspace{-2mm}78.5\hspace{-2mm}&\hspace{-2mm}261.6\hspace{-2mm}&\hspace{-2mm}285.4\hspace{-2mm}&\hspace{-2mm}76.7\hspace{-2mm}&\hspace{-2mm}21.0\hspace{-2mm}&\hspace{-2mm}68\hspace{-2mm}&\hspace{-2mm}1061\hspace{-4mm}& 3847 \hspace{-2mm}& \hspace{-2mm}7.05\hspace{-2mm}&\hspace{-2mm}4.24 \hspace{-2mm}& \hspace{-2mm}3.1\hspace{-2mm}\\

\hspace{-2mm}6\hspace{-2mm}&\hspace{-2mm}3\hspace{-2mm}&\hspace{-2mm}493\hspace{-2mm}&\hspace{-2mm}492\hspace{-2mm}&\hspace{-2mm}493\hspace{-2mm}&\hspace{-2mm}492\hspace{-2mm}&\hspace{-2mm}150.5\hspace{-2mm}&\hspace{-2mm}33.0\hspace{-2mm}&\hspace{-2mm}119.1\hspace{-2mm}&\hspace{-2mm}128.5\hspace{-2mm}&\hspace{-2mm}85.0  \hspace{-2mm}&\hspace{-2mm}9.3\hspace{-2mm}&\hspace{-2mm}82\hspace{-2mm}&\hspace{-2mm}539\hspace{-2mm}& \hspace{-4mm}1738\hspace{-2mm} & \hspace{-2mm}7.05\hspace{-2mm}&\hspace{-2mm}6.50 \hspace{-2mm}& \hspace{-2mm}2.3\hspace{-2mm}\\

  \hspace{-2mm}&\hspace{-2mm}2\hspace{-2mm}&\hspace{-2mm}495\hspace{-2mm}&\hspace{-2mm}494\hspace{-2mm}&\hspace{-2mm}495\hspace{-2mm}&\hspace{-2mm}496\hspace{-2mm}&\hspace{-2mm}221.3\hspace{-2mm}&\hspace{-2mm}16.6\hspace{-2mm}&\hspace{-2mm}56.4 \hspace{-2mm}&\hspace{-2mm}63.3 \hspace{-2mm}&\hspace{-2mm}135.0 \hspace{-2mm}&\hspace{-2mm}1.2\hspace{-2mm}&\hspace{-2mm}46\hspace{-2mm}&\hspace{-2mm}130\hspace{-2mm}& \hspace{-4mm}902\hspace{-2mm} & \hspace{-2mm}7.05\hspace{-2mm}&\hspace{-4mm}6.24\hspace{-2mm} & \hspace{-2mm}2.0\hspace{-2mm}\\
 
\hline \hline
\hspace{-2mm}&\hspace{-2mm}4\hspace{-2mm}&\hspace{-2mm}492\hspace{-2mm}&\hspace{-2mm}491\hspace{-2mm}&\hspace{-2mm}491\hspace{-2mm}&\hspace{-2mm}491\hspace{-2mm}&\hspace{-2mm}1644\hspace{-2mm}&\hspace{-2mm}198.1\hspace{-2mm}&\hspace{-2mm}2776\hspace{-2mm}&\hspace{-2mm}n/a\hspace{-2mm}&\hspace{-2mm}757.0\hspace{-2mm}&\hspace{-2mm}171.3\hspace{-2mm}&\hspace{-2mm}143\hspace{-2mm}&\hspace{-2mm}2953\hspace{-2mm}& \hspace{-5mm}12014\hspace{-5mm} & \hspace{-2mm}65.9\hspace{-2mm}&\hspace{-2mm}23.6 \hspace{-2mm}& \hspace{-2mm}3.5\hspace{-2mm}\\

\hspace{-2mm}7\hspace{-2mm}&\hspace{-2mm}3\hspace{-2mm}&\hspace{-2mm}494\hspace{-2mm}&\hspace{-2mm}494\hspace{-2mm}&\hspace{-2mm}494\hspace{-2mm}&\hspace{-2mm}494\hspace{-2mm}&\hspace{-2mm}1265\hspace{-2mm}&\hspace{-2mm}161.6\hspace{-2mm}&\hspace{-2mm}1809\hspace{-2mm}&\hspace{-2mm}2813\hspace{-2mm}&\hspace{-2mm}718.3\hspace{-2mm}&\hspace{-2mm}143.6\hspace{-2mm}&\hspace{-2mm}112\hspace{-2mm}&\hspace{-2mm}2133\hspace{-2mm}& \hspace{-1mm}7591 \hspace{-2mm}& \hspace{-2mm}65.9\hspace{-2mm}&\hspace{-2mm}51.2 \hspace{-2mm}& \hspace{-2mm}2.4\hspace{-2mm}\\

\hspace{-2mm}&\hspace{-2mm}2\hspace{-2mm}&\hspace{-2mm}494\hspace{-2mm}&\hspace{-2mm}494\hspace{-2mm}&\hspace{-2mm}494\hspace{-2mm}&\hspace{-2mm}494\hspace{-2mm}&\hspace{-2mm}2210\hspace{-2mm}&\hspace{-2mm}27.3\hspace{-2mm}&\hspace{-2mm}317.7&443.6\hspace{-2mm}&\hspace{-2mm}1241\hspace{-2mm}&\hspace{-2mm}12.3\hspace{-2mm}&\hspace{-2mm}101\hspace{-2mm}&\hspace{-2mm}425\hspace{-2mm}& \hspace{-2mm}2109 \hspace{-2mm} & \hspace{-2mm}65.9\hspace{-2mm}&\hspace{-2mm}61.2 \hspace{-2mm}& \hspace{-2mm}2.0\hspace{-2mm}\\

\hline \hline
\hspace{-2mm}&\hspace{-2mm}4\hspace{-2mm}&\hspace{-2mm}n/a\hspace{-2mm}&\hspace{-2mm}490\hspace{-2mm}&\hspace{-2mm}n/a\hspace{-2mm}&\hspace{-2mm}n/a\hspace{-2mm}&\hspace{-2mm}n/a\hspace{-2mm}&\hspace{-2mm}1212\hspace{-2mm}&\hspace{-2mm}n/a\hspace{-2mm}&\hspace{-2mm}n/a\hspace{-2mm}&\hspace{-2mm}n/a\hspace{-2mm}&\hspace{-2mm}804.1\hspace{-2mm}&\hspace{-2mm}n/a\hspace{-2mm}&\hspace{-2mm}8328\hspace{-2mm}& \hspace{-2mm}n/a\hspace{-2mm} & \hspace{-2mm}n/a\hspace{-2mm}&\hspace{-2mm}66.6 \hspace{-2mm}& \hspace{-2mm}4.1\hspace{-2mm}\\

\hspace{-2mm}8\hspace{-2mm}&\hspace{-2mm}3\hspace{-2mm}&\hspace{-2mm}n/a\hspace{-2mm}&\hspace{-2mm}490\hspace{-2mm}&\hspace{-2mm}n/a\hspace{-2mm}&\hspace{-2mm}n/a\hspace{-2mm}&\hspace{-2mm}n/a\hspace{-2mm}&\hspace{-2mm}598.5\hspace{-2mm}&\hspace{-2mm}n/a\hspace{-2mm}&\hspace{-2mm}n/a\hspace{-2mm}&\hspace{-2mm}n/a\hspace{-2mm}&\hspace{-2mm}301.2\hspace{-2mm}&\hspace{-2mm}n/a\hspace{-2mm}&\hspace{-2mm}3956\hspace{-2mm}& \hspace{-2mm}n/a\hspace{-2mm} &\hspace{-2mm}n/a\hspace{-2mm}&\hspace{-2mm}379.7 \hspace{-2mm}& \hspace{-2mm}3.0\hspace{-2mm}\\

\hspace{-2mm}&\hspace{-2mm}2\hspace{-2mm}&\hspace{-2mm}n/a\hspace{-2mm}&\hspace{-2mm}492\hspace{-2mm}&\hspace{-2mm}n/a\hspace{-2mm}&\hspace{-2mm}n/a\hspace{-2mm}&\hspace{-2mm}n/a\hspace{-2mm}&\hspace{-2mm}215.3\hspace{-2mm}&\hspace{-2mm}1908\hspace{-2mm}&\hspace{-2mm}n/a\hspace{-2mm}&\hspace{-2mm}n/a\hspace{-2mm}&\hspace{-2mm}153.2\hspace{-2mm}&\hspace{-2mm}n/a\hspace{-2mm}&\hspace{-2mm}2019\hspace{-2mm}& \hspace{-2mm}7251\hspace{-2mm} & \hspace{-2mm}n/a\hspace{-2mm}&\hspace{-2mm}1121 \hspace{-2mm}& \hspace{-2mm}2.3\hspace{-2mm}\\

\hline \hline
\hspace{-2mm}15\hspace{-2mm}&\hspace{-2mm}3\hspace{-2mm}&\hspace{-2mm}n/a\hspace{-2mm}&\hspace{-2mm}486&\hspace{-2mm}n/a\hspace{-2mm}&\hspace{-2mm}n/a\hspace{-2mm}&\hspace{-2mm}n/a\hspace{-2mm}&\hspace{-2mm}1809\hspace{-2mm}&\hspace{-2mm}n/a\hspace{-2mm}&\hspace{-2mm}n/a\hspace{-2mm}&\hspace{-2mm}n/a\hspace{-2mm}&\hspace{-2mm}1733\hspace{-2mm}&\hspace{-2mm}n/a\hspace{-2mm}&\hspace{-2mm}7276\hspace{-2mm}& \hspace{-2mm}n/a\hspace{-2mm} & \hspace{-2mm}n/a\hspace{-2mm}&\hspace{-2mm}$1.2\cdot 10^{7}$\hspace{-2mm} & \hspace{-2mm}5.7\hspace{-2mm}\\

\hline \hline

\hspace{-2mm}17\hspace{-2mm}&\hspace{-2mm}4\hspace{-2mm}&\hspace{-2mm}n/a\hspace{-2mm}&\hspace{-2mm}481\hspace{-2mm}&\hspace{-2mm}n/a\hspace{-2mm}&\hspace{-2mm}n/a\hspace{-2mm}&\hspace{-2mm}n/a\hspace{-2mm}&\hspace{-2mm}3548\hspace{-2mm}&\hspace{-2mm}n/a\hspace{-2mm}&\hspace{-2mm}n/a\hspace{-2mm}&\hspace{-2mm}n/a\hspace{-2mm}&\hspace{-2mm}1751\hspace{-2mm}&\hspace{-2mm}n/a\hspace{-2mm}&\hspace{-2mm}15225\hspace{-2mm}& \hspace{-2mm}n/a\hspace{-2mm} & \hspace{-2mm}n/a\hspace{-2mm}&\hspace{-2mm}$2.5\cdot 10^{7}$\hspace{-2mm} & \hspace{-2mm}6.1\hspace{-2mm}\\

\hline

\end{tabular}
\end{center}

 \caption{Performance comparison of $\flucap$ (denoted as \textsf{FluCaP}) and symbolic $\LAO$ (denoted as \textsf{LAO*}), where the cells n/a denote the fact that a planner did not deliver a solution within the time limit of one hour. \textsf{NAS} and \textsf{NGS} are number of abstract and ground states, respectively.}
\label{t:2}
\end{table*}

At the outset of solving a colored BW problem, symbolic $\LAO$ starts by propositionalizing its components, namely, the goal statement and actions. Only after that, the abstraction using propositional ADDs is applied. In contrast, $\flucap$ performs first-order abstraction on a colored BW problem directly, avoiding unnecessary grounding. In the following, we show how an abstraction technique affects the computation of a heuristic function. To create an admissible heuristic, $\flucap$ performs twenty iterations of $\FOVI$ and symbolic $\LAO$ performs twenty iterations of an approximate value iteration algorithm similar to APRICODD by~\citeA{staubin}. The columns labelled \textsf{H.time} and \textsf{NAS} show the time needed for computing a heuristic function and the number of abstract states it covers, respectively. In comparison to $\flucap$, symbolic $\LAO$ needs to evaluate fewer abstract states in the heuristic function but takes considerably more time. One can conclude that abstract states in symbolic $\LAO$ enjoy more complex structure than those in $\flucap$.

We note that, in comparison to $\FOVI$, $\flucap$ restricts the value iteration to a smaller state space. Intuitively, the value function, which is delivered by $\FOVI$, covers a larger state space, because the time that is allocated for the heuristic search in $\flucap$ is now used for performing additional iterations in $\FOVI$. The results in the column labelled \textsf{\%} justify that the harder the problem is (that is, the more colors it contains), the higher the percentage of runtime spent on normalization. Almost on all test problems, the effort spent on normalization takes three percent of the total runtime on average.

In order to compare the heuristic accuracy, we present in the column labelled \textsf{NGS} the number of ground states which the heuristic assigns non-zero values to. One can see that the heuristics returned by $\flucap$ and symbolic $\LAO$ have similar accuracy, but $\flucap$ takes much less time to compute them. This reflects the advantage of the plain first-order abstraction in comparison to the marriage of propositionalization with abstraction using propositional ADDs. In some examples, we gain several orders of magnitude in \textsf{H.time}.

The column labelled \textsf{Total time} presents the time needed to solve a problem. During this time, a planner must execute 30 runs from an initial state to a goal state. A one-hour block is allocated for each problem. We note that, in comparison to $\flucap$, the time required by heuristic search in symbolic $\LAO$ (i.e., difference between \textsf{Total time} and \textsf{H.time}) grows considerably faster in the size of the problem. This reflects the potential of employing first-order abstraction instead of abstraction based on propositional ADDs during heuristic search.

The average reward obtained over 30 runs, shown in column \textsf{Total av.\ reward}, is the planner's evaluation score. The reward value close to 500 (which is the maximum possible reward) simply indicates that a planner found a reasonably good policy. Each time the number of blocks \textsf{B} increases by 1, the running time for symbolic $\LAO$ increases roughly 10 times. Thus, it could not scale to problems having more than seven blocks. This is in contrast to $\flucap$ which could solve problems of seventeen blocks. We note that the number of colors \textsf{C} in a problem affects the efficiency of an abstraction technique. In $\flucap$, as \textsf{C} decreases, the abstraction rate increases which, in turn, is reflected by the dramatic decrease in runtime. The opposite holds for symbolic $\LAO$.

In addition, we compare $\flucap$ with two variants. The first one, denoted as \textsf{FOVI}, performs no heuristic search at all, but rather, employs $\FOVI$ to compute the $\varepsilon$-optimal total value function from which a policy is extracted. The second one, denoted as \textsf{FluCaP}$^\textsf{--}$, performs `trivial' heuristic search starting with an initial value function as an admissible heuristic. As expected, $\flucap$ that combines heuristic search and $\FOVI$ demonstrates an advantage over plain $\FOVI$ and trivial heuristic search. These results illustrate the significance of heuristic search in general (\textsf{FluCaP} vs.\ \textsf{FOVI}) and the importance of heuristic accuracy, in particular (\textsf{FluCaP} vs.\ \textsf{FluCaP}$^\textsf{--}$). \textsf{FOVI} and \textsf{FluCaP}$^\textsf{--}$ do not scale to problems with more than seven blocks.

\begin{table*}[tb]

  \footnotesize
\begin{center}  
\begin{tabular}{|c||c||c||c||c||c|}
   \hline
\textsf{B} & 
\textsf{Total av. reward, $\leq$500} &
\textsf{Total time, sec.} & 
\textsf{H.time, sec.} & 
\textsf{NAS} & 
\textsf{NGS $\times\ \textsf{10}^\textsf{21}$}\\ 

\hline \hline

20 & 489.0 & 137.5 & 56.8 & 711 & 1.7\\
22 & 487.4 & 293.8 & 110.2 & 976 &  1.1 $\times\ \textsf{10}^\textsf{3}$\\
24 & 492.0 & 757.3 & 409.8 & 1676 &  1.0 $\times\ \textsf{10}^\textsf{6}$\\
26 & 482.8 & 817.0 & 117.2 &  1141 &  4.6 $\times\ \textsf{10}^\textsf{8}$\\
28 & 493.0 & 2511.3 & 823.3 &  2832 & 8.6 $\times\ \textsf{10}^\textsf{11}$\\
30 & 491.2 & 3580.4 & 1174.0 &  4290 & 1.1  $\times\ \textsf{10}^\textsf{15}$ \\
32 & 476.0 & 3953.8 & 781.8 &  2811 & 7.4  $\times\ \textsf{10}^\textsf{17}$\\
34 & 475.6 & 3954.1 & 939.4 &  3248 &  9.6$\times\ \textsf{10}^\textsf{20}$\\
36 & n/a & n/a &  n/a &  n/a &  n/a\\
\hline

\end{tabular}
\end{center}

 \caption{Performance of $\flucap$ on larger instances of one-color Blocksworld problems, where the cells n/a denote the fact that a planner did not deliver a solution within the time limit.}
 \label{t:3}
\end{table*}

Table~\ref{t:3} presents the performance results of $\flucap$ on larger instances of one-color BW problems with the number of blocks varying from twenty to thirty four. We believe that $\flucap$ does not scale to problems of larger size because the implementation is not yet well optimized. In general, we believe that the $\flucap$ system should not be as sensitive to the size of a problem as propositional planners are.

Our experiments were targeted at the one-color problems only because they are, on the one hand, the simplest ones for us and, on the other hand, the bottleneck for propositional planners. The structure of one-color problems allows us to apply first-order state abstraction in its full power. For example, for a 34-blocks problem $\flucap$ operates on about 3.3 thousand abstract states that explode to $9.6\times 10^{41}$ individual states after propositionalization. A propositional planner must be highly optimized in order to cope with this non-trivial state space.

We note that additional colors in larger instances (more than 20 blocks) of BW problems cause dramatic increase in computational time, so we consider these problems as being unsolved. One should also observe that the number of abstract states \textsf{NAS} increases with the number of blocks non-monotonically because the problems are generated randomly. For example, the 30-blocks problem happens to be harder than the 34-blocks one. Finally, we note that all results that appear in Tables~\ref{t:2} and~\ref{t:3} were obtained by using the new version of the evaluation software that does not rely on propositionalization in contrast to the initial version that was used during the competition.

Table~\ref{t:4} presents the competition results from IPC'2004, where $\flucap$ was competitive in comparison with other planners on colored BW problems. $\flucap$ did not perform well on non-colored BW problems because these problems were propositional ones (that is, goal statements and initial states are ground) and $\flucap$ does not yet incorporate optimization techniques applied in modern propositional planners.  The contestants are indicated by their origin. For example, Dresden - $\flucap$, UMass - symbolic $\LAO$ etc. Because only the $\pickup$ action has cost 1, the gain of five points in total reward means that the plan contains ten fewer actions on average. The competition domains and log files are available in an online appendix of~\citeA{younes}.

Although the empirical results that are presented in this work were obtained on the domain-dependent version of $\flucap$, we have recently developed in~\cite{karabaev} an efficient domain-independent inference mechanism that is the core of a domain-independent version of $\flucap$.

\begin{table*}[tb]

  \footnotesize
\begin{center}  
\begin{tabular}{|c|c||c|c|c|c|c|c|c|c|c|}
   \hline
\multicolumn{2}{|c||}{\textsf{Problem}} &
\multicolumn{9}{c|}{\textsf{Total av. reward, $\leq$500}} 
\\ \cline{1-11}
 \textsf{B} & \textsf{C} & \rotatebox{90}{\textsf{Canberra}} & \rotatebox{90}{\textsf{\textbf{Dresden}}} & \rotatebox{90}{\textsf{UMass}} & \rotatebox{90}{\textsf{Michigan}} & \rotatebox{90}{\textsf{Purdue$_\textsf{1}$}} & \rotatebox{90}{\textsf{Purdue$_\textsf{2}$}} & \rotatebox{90}{\textsf{Purdue$_\textsf{3}$}} & \rotatebox{90}{\textsf{Caracas}} & \rotatebox{90}{\textsf{Toulouse}}
\\
\hline \hline

 5  & 3 & 494.6 & 496.4 & n/a & n/a & 496.5 & 496.5 & 495.8 & n/a & n/a \\ 

 8  & 3 & 486.5 & 492.8 & n/a & n/a & 486.6 & 486.4 & 487.2 & n/a & n/a\\

11 & 5 & 479.7 & 486.3 & n/a & n/a & 481.3 & 481.5 & 481.9 & n/a & n/a \\

\hline \hline
5 & 0 & 494.6 & 494.6 & 494.8 & n/a & 494.1 & 494.6 & 494.4 & 494.9 & 494.1 \\
8 & 0 & 489.7 & 489.9 & n/a & n/a & 488.7 & 490.3 & 490 & 488.8 & n/a\\
11 & 0 & 479.1 & n/a & n/a & n/a & 480.3 & 479.7 & 481.1 & 465.7 & n/a\\
15 & 0 & 467.5 & n/a & n/a & n/a & 469.4 & 467.7 & 486.3 & 397.2 & n/a \\
18 & 0 & 351.8 & n/a & n/a & n/a & 462.4 & -54.9 & n/a & n/a & n/a\\
21 & 0 & 285.7 & n/a & n/a & n/a & 455.7 & 455.1 & 459 & n/a & n/a\\

\hline

\end{tabular}
\end{center}

\caption{Official competition results for colored and non-colored Blocksworld scenarios. May, 2004. The n/a-entries in the table indicate that either a planner was not successful in solving a problem or did not attempt to solve it. }
\label{t:4}
\end{table*}

\section{Related Work}

We follow the symbolic DP (SDP) approach within Situation Calculus (SC) of~\shortciteA{boutilierB} in using first-order state abstraction for FOMDPs. One difference is in the representation language: We use PFC instead of SC. In the course of symbolic value iteration, a state space may contain redundant abstract states that dramatically affect the algorithm's efficiency. In order to achieve computational savings, normalization must be performed to remove this redundancy. However, in the original work by~\shortciteA{boutilierB} this was done by hand. To the best of our knowledge, the preliminary implementation of the SDP approach within SC uses human-provided rewrite rules for logical simplification. In contrast,~\shortciteA{hoelldoblerA} have developed an automated normalization procedure for FOVI that is incorporated in the competition version of $\flucap$ and brings the computational gain of several orders of magnitude. Another crucial difference is that our algorithm uses heuristic search to limit the number of states for which a policy is computed. 

The ReBel algorithm by~\citeA{kersting} relates to $\FOLAO$ in that it also uses a representation language that is simpler than Situation Calculus. This feature makes the state space normalization computationally feasible.

In motivation, our approach is closely connected to Relational Envelope-based Planning (REBP) by~\shortciteA{gardiol} that represents MDP dynamics by a compact set of relational rules and extends the envelope method by~\shortciteA{dean}. However, REBP propositionalizes actions first, and only afterwards employs abstraction using equivalence-class sampling. In contrast, $\FOLAO$ directly applies state and action abstraction on the first-order structure of an MDP. In this respect, REBP is closer to symbolic $\LAO$ than to $\FOLAO$. Moreover, in contrast to PFC, action descriptions in REBP do not allow negation to appear in preconditions or in effects. In organization, $\FOLAO$, as symbolic $\LAO$, is similar to real-time DP by~\shortciteA{barto} that is an online search algorithm for MDPs. In contrast, $\FOLAO$ works offline.

All the above algorithms can be classified as deductive approaches to solving FOMDPs. They can be characterized by the following features: (1) they are model-based, (2) they aim at exact solutions, and (3) logical reasoning methods are used to compute abstractions. We should note that FOVI aims at exact solution for a FOMDP, whereas $\FOLAO$, due to the heuristic search that avoids evaluating all states, seeks for an approximate solution. Therefore, it would be more appropriate to classify $\FOLAO$ as an approximate deductive approach to FOMDPs.

In another vein, there is some research on developing inductive approaches to solving FOMDPs, e.g., by~\citeA{fern}. The authors propose the approximate policy iteration (API) algorithm, where they replace the use of cost-function approximations as policy representations in API with direct, compact state-action mappings, and use a standard relational learner to learn these mappings. In effect,~\citeauthor{fern} provide policy-language biases that enable solution of very large relational MDPs. All inductive approaches can be characterized by the following features: (1) they are model-free, (2) they aim at approximate solutions, and (3) an abstract model is used to generate biased samples from the underlying FOMDP and the abstract model is altered based on them.

A recent approach by~\shortciteA{grettonB} proposes an inductive policy construction algorithm that strikes a middle-ground between deductive and inductive techniques. The idea is to use reasoning, in particular first-order regression, to automatically generate a hypothesis language, which is then used as input by an inductive solver. The approach by~\citeauthor{grettonB} is related to SDP and to our approach in the sense that a first-order domain specification language as well as logical reasoning are employed.

\section{Conclusions}

We have proposed an approach that combines heuristic search and first-order state abstraction for solving FOMDPs more efficiently. Our approach can be seen as two-fold: First, we use dynamic programming to compute an approximate value function that serves as an admissible heuristic. Then heuristic search is performed to find an exact solution for those states that are reachable from the initial state. In both phases, we exploit the power of first-order state abstraction in order to avoid evaluating states individually. As experimental results show, our approach breaks new ground in exploring the efficiency of first-order representations in solving MDPs. In comparison to existing MDP planners that must propositionalize the domain, e.g., symbolic $\LAO$, our solution scales better on larger FOMDPs.

However, there is plenty remaining to be done. For example, we are interested in the question of to what extent the optimization techniques applied in modern propositional planners can be combined with first-order state abstraction. In future competitions, we would like to face problems where the goal and/or initial states are only partially defined and where the underlying domain contains infinitely many objects.

The current version of $\FOLAO$ is targeted at the problems that allow for efficient first-order state abstraction. More precisely, these are the problems that can be polynomially translated into PFC. For example in the colored BW domain, existentially-closed goal descriptions were linearly translated into the equivalent PFC representation. Whereas universally-closed goal descriptions would require full propositionalization. Thus, the current version of PFC is less first-order expressive than, e.g., Situation Calculus. In the future, it would be interesting to study the extensions of the PFC language, in particular, to find the trade-off between the PFC's expressive power and the tractability of solution methods for FOMDPs based on PFC.

\subsubsection*{Acknowledgements} 
We are very grateful to all anonymous reviewers for the thorough reading of the previous versions of this paper. We also thank Zhengzhu Feng for fruitful discussions and for providing us with the executable of the symbolic $\LAO$ planner. We greatly appreciate David E.\ Smith for his patience and encouragement. His valuable comments have helped us to improve this paper. Olga Skvortsova was supported by a grant within the Graduate Programme GRK 334 ``Specification of discrete processes and systems of processes by operational models and logics" under auspices of the Deutsche Forschungsgemeinschaft (DFG).


\bibliography{all_new}
\bibliographystyle{theapa}

\end{document}